\documentclass{article}

    \usepackage[preprint,nonatbib]{neurips_2020}

\usepackage[utf8]{inputenc} 
\usepackage[T1]{fontenc}    
\usepackage{hyperref}       
\usepackage{url}            
\usepackage{booktabs}       
\usepackage{amsfonts}       
\usepackage{nicefrac}       
\usepackage{microtype}      
\usepackage{graphicx}
\usepackage{subfigure}
\usepackage{booktabs} 
\usepackage{amsfonts}
\usepackage{amsmath}
\usepackage{amssymb}
\usepackage[numbers]{natbib}
\usepackage{xcolor}
\usepackage[ruled,vlined]{algorithm2e}
\usepackage{algorithmic}
\usepackage{wrapfig}
\usepackage{caption}
\usepackage{wrapfig}
\allowdisplaybreaks

\title{Deep Reinforcement Learning\\amidst Lifelong Non-Stationarity}

\author{%
  Annie Xie, James Harrison, Chelsea Finn\\
  Stanford University, Stanford, CA\\
  \texttt{\{anniexie,jharrison,cbfinn\}@stanford.edu} 
}

\begin{document}

\maketitle

\begin{abstract}
As humans, our goals and our environment are persistently changing throughout our lifetime based on our experiences, actions, and internal and external drives. In contrast, typical reinforcement learning problem set-ups consider decision processes that are stationary across episodes. Can we develop reinforcement learning algorithms that can cope with the persistent change in the former,  more realistic problem settings? While on-policy algorithms such as policy gradients in principle can be extended to non-stationary settings, the same cannot be said for more efficient off-policy algorithms that replay past experiences when learning. In this work, we formalize this problem setting, and draw upon ideas from the online learning and probabilistic inference literature to derive an off-policy RL algorithm that can reason about and tackle such lifelong non-stationarity. Our method leverages latent variable models to learn a representation of the environment  from current and past experiences, and performs off-policy RL with this representation. We further introduce several simulation environments that exhibit lifelong non-stationarity, and empirically find that our approach substantially outperforms approaches that do not reason about environment shift.\footnote{Videos of our results are available at \url{https://sites.google.com/stanford.edu/lilac/}}

\end{abstract}

\section{Introduction}
\label{sec:introduction}

In the standard reinforcement learning (RL) set-up, the agent is assumed to operate in a stationary environment, i.e., under fixed dynamics and reward. However, the assumption of stationarity rarely holds in more realistic settings, such as in the context of lifelong learning systems~\cite{thrun1998lifelong}. That is, over the course of its lifetime, an agent may be subjected to environment dynamics and rewards that vary with time. In robotics applications, for example, this non-stationarity manifests itself in changing terrains and weather conditions. In some situations, not even the objective is necessarily fixed: consider an assistive robot helping a human whose preferences gradually change over time. And, because stationarity is a core assumption in many existing RL algorithms, they are unlikely to perform well in these environments.

Crucially, in each of the above scenarios, the environment is specified by unknown, time-varying parameters. These latent parameters are also not i.i.d., e.g., if the sky is clear at this very moment, it likely will not suddenly start raining in the next; in other words, these parameters have associated but unobserved dynamics. In this paper, we formalize this problem setting with the dynamic parameter Markov decision process (DP-MDP). The DP-MDP corresponds to a sequence of stationary MDPs, related through a set of latent parameters governed by an autonomous dynamical system. While all non-stationary MDPs are special instances of the partially observable Markov decision process (POMDP)~\cite{kaelbling1998planning}, in this setting, we can leverage structure available in the dynamics of the hidden parameters and avoid solving POMDPs in the general case.

On-policy RL algorithms can in principle cope with such non-stationarity~\cite{sutton2007role}. However, in highly dynamic environments, only a limited amount of interaction is permitted before the environment changes, and on-policy methods may fail to adapt rapidly enough in this low-shot setting~\cite{al2017continuous}. Instead, we desire an off-policy RL algorithm that can use past experience both to improve sample efficiency and to reason about the environment dynamics. In order to adapt, the agent needs the ability to predict how the MDP parameters will shift. We thus require a representation of the MDP as well as a model of how parameters evolve in this space, both of which can be learned from off-policy experience.

To this end, our core contribution is an off-policy RL algorithm that can operate under non-stationarity by jointly learning (1) a latent variable model, which lends a compact representation of the MDP, and (2) a maximum entropy policy with this representation. We validate our approach, which we call \textbf{Li}felong \textbf{L}atent \textbf{A}ctor-\textbf{C}ritic (LILAC), on a set of simulated environments that demonstrate persistent non-stationarity. In our experimental evaluation, we find that our method far outperforms RL algorithms that do not account for environment dynamics.

\section{Dynamic Parameter Markov Decision Processes}
\label{sec:problem_statement}

\newcommand{\ssp}{\mathcal{S}}
\newcommand{\asp}{\mathcal{A}}

\newcommand{\at}{\mathbf{a}_t}

\newcommand{\s}{\mathbf{s}}
\newcommand{\sone}{\mathbf{s}_1}
\newcommand{\st}{\mathbf{s}_t}
\newcommand{\stp}{\mathbf{s}_{t+1}}

\newcommand{\zsp}{\mathcal{Z}}
\newcommand{\z}{\mathbf{z}}

\newcommand{\opt}{\mathcal{O}}
\newcommand{\optt}{\opt_t}
\newcommand{\ac}{\mathbf{a}}

The standard RL setting assumes episodic interaction with a fixed MDP~\cite{sutton2018reinforcement}. In the real world, the assumption of episodic interaction with identical MDPs is limiting as it does not capture the wide variety of exogenous factors that may effect the decision-making problem. A common model to avoid the strict assumption of Markovian observations is the partially observed MDP (POMDP) formulation~\cite{kaelbling1998planning}. While the POMDP is highly general, we focus in this work on leveraging known structure of the non-stationary MDP to improve performance. In particular, we consider an episodic environment, which we call the \textit{dynamic parameter MDP} (DP-MDP), where a new MDP (we also refer to MDPs as tasks) is presented in each episode. In reflection of the regularity of real-world non-stationarity, the tasks are sequentially related through a set of continuous parameters.

Formally, the DP-MDP is equipped with state space $\ssp$, action space $\asp$, and initial state distribution $\rho_\s(\sone)$. 
Following the formulation of the Hidden Parameter MDP (HiP-MDP)~\cite{doshi2016hidden},
a set of \textit{unobserved} task parameters $\z \in \zsp$ defines the dynamics $p_\s(\stp | \st, \at; \z)$ and reward function $r(\st, \at ; \z)$ for each task. 
In contrast to the HiP-MDP, the task parameters $\z$ in the DP-MDP are not sampled i.i.d. but instead shift stochastically according to $p_\z(\z^{i+1} | \z^i)$, with initial distribution $\rho_\z (\z^1)$. In other words, the DP-MDP is a sequence of tasks with parameters determined by the transition function $p_\z$. If the task parameters $\z$ for each episode were known, the augmented state space $\ssp \times \zsp$ would define a fully observable MDP for which we can use standard RL algorithms. Hence, in our approach, we aim to infer the hidden task parameters and learn their transition function, allowing us to leverage existing RL algorithms by augmenting the observations with the inferred task parameters.

\section{Preliminaries: RL as Inference}
\label{sec:background}

We first discuss an established connection between probabilistic inference and reinforcement learning~\cite{toussaint2009robot,levine2018reinforcement} to provide some context for our approach. At a high level, this framework casts sequential decision-making as a probabilistic graphical model, and from this perspective, the maximum-entropy RL objective can be derived as an inference procedure in this model.

\newcommand{\zone}{\z^1}
\newcommand{\zi}{\z^i}
\newcommand{\zim}{\z^{i-1}}

\subsection{A Probabilistic Graphical Model for RL}

\begin{figure}
    \centering
    \begin{minipage}{.4\textwidth}
        \centering
        \vspace{0.4cm}
        \includegraphics[width=.8\linewidth]{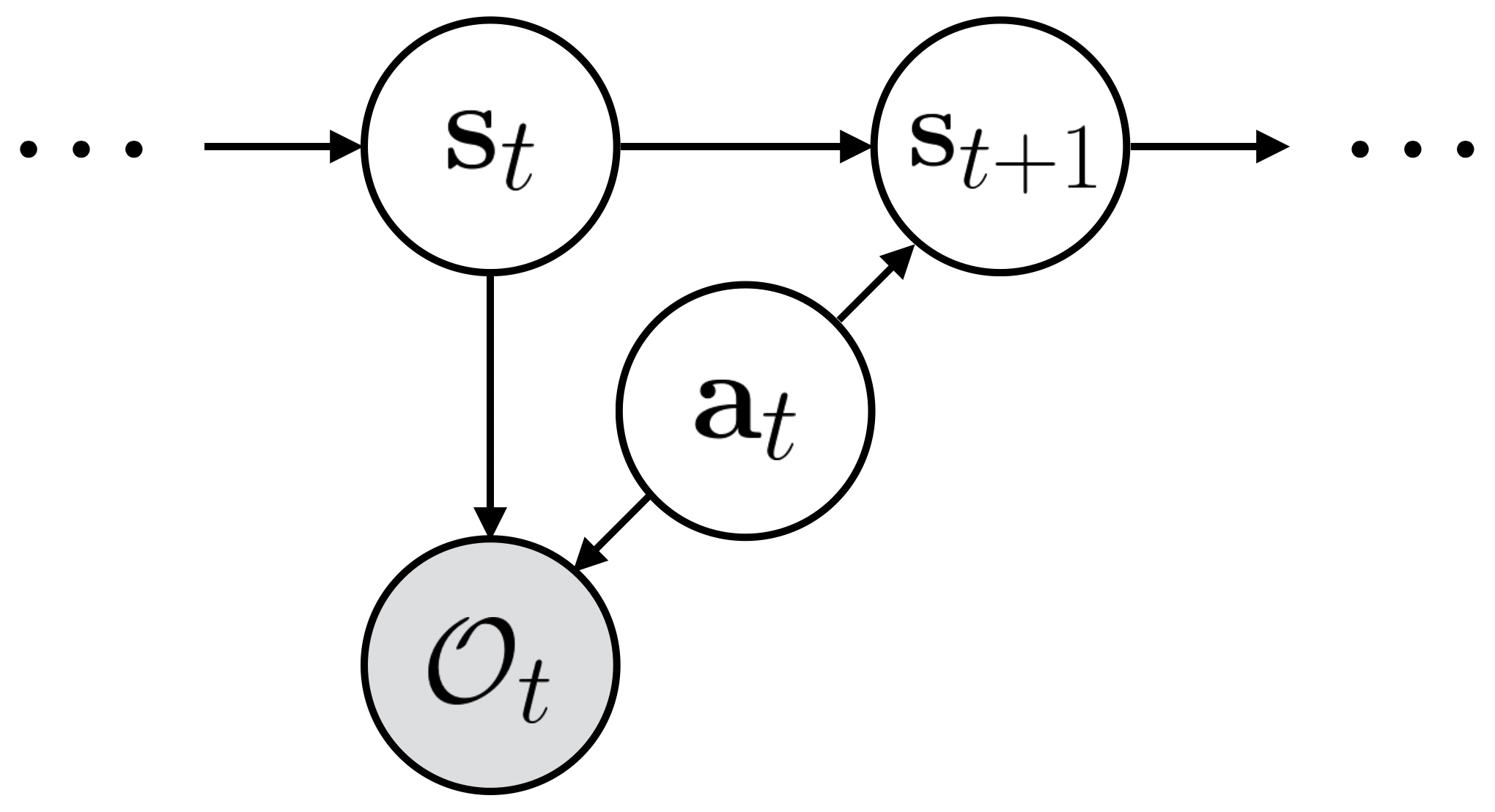} 
        \vspace{0.25cm}
        \captionof{figure}{\footnotesize The graphical model for the RL-as-Inference framework consists of states $\st$, actions $\at$, and optimality variables $\opt_t$. By incorporating rewards through the optimality variables, learning an RL policy amounts to performing inference in this model.}
        \label{fig:control_as_inference}
    \end{minipage}
    \hspace{0.05\linewidth}
    \begin{minipage}{.52\textwidth}
        \centering
        \includegraphics[width=\linewidth]{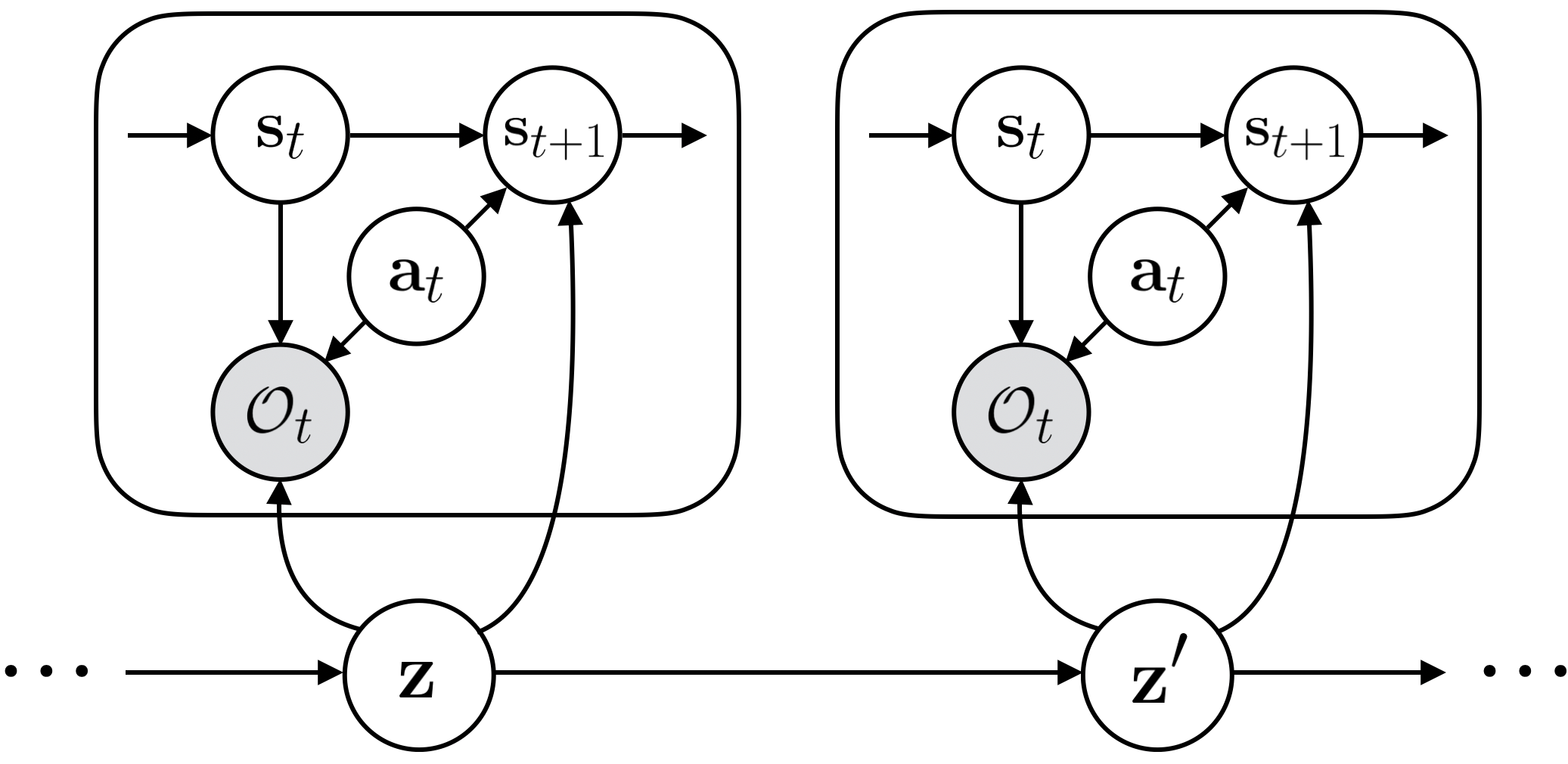}
        \captionof{figure}{\footnotesize The graphical model for the DP-MDP. Each episode presents a new task, or MDP, determined by latent variables $\z$. The MDPs are further sequentially related through a transition function $p_\z(\z' | \z)$.}
        \label{fig:dpmdp_as_inference}
    \end{minipage}
\end{figure}

As depicted in Figure~\ref{fig:control_as_inference}, the proposed model consists of states $\st$, actions $\at$, and per-timestep optimality variables $\opt_t$, which are related to rewards by $p(\opt_t = 1 | \st, \at) = \exp(r(\st, \at))$ and denote whether the action $\at$ taken from state $\st$ is optimal. While rewards are required to be non-positive through this relation, so long the rewards are bounded, they can be scaled and centered to be no greater than 0.
A trajectory is the sequence of states and actions, $(\s_1, \ac_1, \s_2, \dots, \s_T, \ac_T)$, and we aim to infer the posterior distribution $p(\s_{1:T}, \ac_{1:T} | \opt_{1:T} = 1)$, i.e., the trajectory distribution that is optimal for all timesteps.

\subsection{Variational Inference}
\label{subsection:variational_inference}
Among existing inference tools, structured variational inference is particularly appealing for its scalability and efficiency to approximate the distribution of interest. In the variational inference framework, a variational distribution $q$ is optimized through the variational lower bound to approximate another distribution $p$. Assuming a uniform prior over actions, the optimal trajectory distribution is:
$$
p(\s_{1:T}, \ac_{1:T} | \opt_{1:T} = 1) \propto p(\s_{1:T}, \ac_{1:T}, \opt_{1:T} = 1) = p(\sone) \prod_{t=1}^T \exp(r(\st, \at)) p(\stp | \st, \at).
$$
We can choose the form
$
q(\s_{1:T}, \ac_{1:T}) = p(\sone) \prod_{t=1}^T p(\stp | \st, \at) q(\at | \st)
$ for our approximating distribution,
where $p(\sone)$ and $p(\stp | \st, \at)$ are fixed and given by the environment. We now rename $q(\at | \st)$ to $\pi(\at | \st)$ since this represents the desired policy. By Jensen's inequality, the variational lower bound for the evidence $\opt_{1:T} = 1$ is given by
\begin{align*}
    \log p(\opt_{1:T} = 1) &= \log \mathbb{E}_{q} \left[ \frac{p(\s_{1:T}, \ac_{1:T}, \opt_{1:T} = 1)}{q(\s_{1:T}, \ac_{1:T})} \right] \ge \mathbb{E}_{\pi} \left[ \sum_{t=1}^T r(\st, \at) - \log \pi(\at | \st) \right],
\end{align*}
which is the maximum entropy RL objective~\cite{ziebart2008maximum,toussaint2009robot,rawlik2013stochastic,fox2015taming,haarnoja2017reinforcement}. This objective adds a conditional entropy term and thus maximizes both returns and the entropy of the policy. This formulation is known for its improvements in exploration, robustness, and stability over other RL algorithms, thus we build upon it in our method to inherit these qualities. We capture non-stationarity by augmenting the RL-as-inference model with latent variables $\textbf{z}^i$ for each task $i$. As we will see in the next section, by viewing non-stationarity from this probabilistic perspective, our algorithm can be derived as an inference procedure in a unified model.

\section{Off-Policy Reinforcement Learning in Non-Stationary Environments}
\label{sec:method}
Building upon the RL-as-inference framework, in this section, we offer a probabilistic graphical model that underlies the dynamic parameter MDP setting introduced in Section~\ref{sec:problem_statement}. Then, using tools from variational inference, we derive a variational lower bound that performs joint RL and representation learning. Finally, we present our RL algorithm, which we call \textbf{Li}felong \textbf{L}atent \textbf{A}ctor-\textbf{C}ritic (LILAC), that optimizes this objective and builds upon on soft actor-critic~\cite{haarnoja2018soft}, an off-policy maximum entropy RL algorithm.

\subsection{Non-stationarity as a Probabilistic Model}
We can cast the dynamic parameter MDP as a probabilistic hierarchical model, where non-stationarity occurs at the episodic level, and within each episode is an instance of a stationary MDP. To do so, we construct a two-tiered model: on the first level, we have the sequence of latent variables $\zi$ as a Markov chain, and on the second level, a Markov decision process corresponding to each $\zi$. The graphical model formulation of the DP-MDP is illustrated in Figure~\ref{fig:dpmdp_as_inference}.

Within this formulation, the trajectories gathered from each episode are modeled individually, rather than amortized as in Subsection~\ref{subsection:variational_inference}, and the joint probability distribution is defined as follows:
\begin{align*}
    p(\z^{1:N}, \tau^{1:N}) &= p(\zone) p(\tau^1 | \zone) \prod_{i=1}^N p(\zi | \zim) p(\tau^i | \zi)
\end{align*}
where the probability of each trajectory $\tau$ given $\z$, assuming a uniform prior over actions, is
\begin{align*}
    p(\tau | \z) &= p(\sone) \prod_{t=1}^T p(\optt = 1 | \st, \at; \z) p(\stp | \st, \at; \z) = p(\sone) \prod_{t=1}^T \exp(r(\st, \at; \z)) p(\stp | \st, \at; \z).
\end{align*}
With this factorization, the non-stationary elements of the environment are captured by the latent variables $\z$, and within a task, the dynamics and reward functions are necessarily stationary. This suggests that learning to infer $\z$, which amounts to representing the non-stationarity elements of the environment with $\z$, will reduce this RL setting to a stationary one. Taking this type of approach is appealing since there already exists a rich body of algorithms for the standard RL setting. In the next subsection, we describe how we can approximate the posterior over $\z$, by deriving the evidence lower bound for this model under the variational inference framework. 

\subsection{Joint Representation and Reinforcement Learning via Variational Inference}
\label{subsection:objective}

Recall the agent is operating in an online learning setting. That is, it must continuously adapt to a stream of tasks and leverage experience gathered from previous tasks for learning. Thus, at any episode $i > 1$, the agent has observed all of the trajectories collected from episodes $1$ through $i-1$, $\tau^{1:i-1} = \{ \tau^1, \cdots, \tau^{i-1} \}$, where $\tau = \{ \s_1, \ac_1, r_1, \dots, \s_T, \ac_T, r_T \}$.

We aim to infer, at every episode $i$, the posterior distribution over actions, given the evidence $\opt_{1:T}^i = 1$ and the experience from the previous episodes $\tau^{1:i-1}$. Following Subsection~\ref{subsection:variational_inference}, we can leverage variational inference to optimize a variational lower bound to the log-probability of this set of evidence, $\log p(\tau^{1:i-1}, \opt_{1:T}^i = 1)$. Since $p(\tau^{1:i-1}, \opt_{1:T}^i = 1)$ factorizes as $p(\tau^{1:i-1}) p(\opt_{1:T}^i = 1 | \tau^{1:i-1})$, the log-probability of the evidence can be decomposed into $\log p(\tau^{1:i-1}) + \log p(\opt_{1:T}^i = 1 | \tau^{1:i-1})$. These two terms can be separately lower bounded and summed to form a single objective.

The variational lower bound of the first term follows from that of a variational auto-encoder~\cite{kingma2013auto} with evidence $\tau^{1:i-1}$ and latent variables $\z^{1:i-1}$:
\begin{align*}
    \log p(\tau^{1:i-1}) 
    &= \log \mathbb{E}_{q} \left[ \frac{p(\tau^{1:i-1}, \z^{1:i-1})}{q(\z^{1:i-1})} \right].
\end{align*}
We choose our approximating distribution over the latent variables $\zi$ to be conditioned on the trajectory from episode $i$, i.e. $q(\zi | \tau^i)$. Then, the variational lower bound can be expressed as:
\begin{align*}
    \log p(\tau^{1:i-1}) &\ge \mathbb{E}_{q} \left[ \sum_{i'=1}^i \sum_{t=1}^T \log p(\stp, r_t | \st, \at; \z^{i'}) - D_\text{KL} (q(\z^{i'} | \tau^{i'})) ~||~ p(\z^{i'} | \z^{i'-1})) \right] = \mathcal{L}_\text{rep}.
\end{align*}
The lower bound $\mathcal{L}_\text{rep}$ corresponds to an objective for unsupervised representation learning in a sequential latent variable model. By optimizing the reconstruction loss of the transitions and rewards for each episode, the learned latent variables should encode the varying parameters of the MDP. Further, by imposing the prior $p(\zi | \zim)$ on the approximated distribution $q$ through the KL divergence, the latent variables are encouraged to be sequentially consistent across time. This prior corresponds to a model of the environment's latent dynamics and gives the agent a predictive estimate of future conditions of the environment (to the extent to which the DP-MDP is predictable).

For the second term,
\begin{align*}
    \log p(\opt_{1:T}^i = 1 | \tau^{1:i-1}) &= \log \int p(\opt_{1:T}^i=1, \zi | \tau^{1:i-1}) d\zi \\
    &= \log \int  p(\opt_{1:T}^i=1 | \zi) p(\zi | \tau^{1:i-1}) d\zi \\
    &\ge \mathbb{E}_{p(\zi | \tau^{1:i-1})} \left[ \log p(\opt^i_{1:T} = 1 | \zi) \right] \\
    &\ge \mathop{\mathbb{E}}_{\substack{p(\zi | \tau^{1:i-1}) \\ \pi(\at | \st, \zi)}} \left[ \sum_{i=1}^T r(\st, \at; \zi) - \log \pi(\at | \st, \zi)  \right] = \mathcal{L}_\text{RL}.
\end{align*}
The final inequality is given by steps from Subsection~\ref{subsection:variational_inference}.
The bound $\mathcal{L}_\text{RL}$ optimizes for both policy returns and policy entropy, as in the maximum entropy RL objective, but here the policy is also conditioned on the inferred latent embeddings of the MDP. This objective essentially performs task-conditioned reinforcement learning where the task variables at episode $i$ are given by $p(\zi | \tau^{1:i-1})$. Learning a multi-task RL policy is appealing, especially over a policy that adapts between episodes. That is, if the shifts in the environment are similar to those seen previously, we do not expect its performance to degrade even if the environment is shifting quickly, whereas a single-task policy would likely struggle to adapt quickly enough.

Our proposed objective is the sum of the above two terms $\mathcal{L} = \mathcal{L}_\text{rep} + \mathcal{L}_\text{RL}$,
which is also a variational lower bound for our entire model. Hence, while our objective was derived from and can be understood as an inference procedure in our probabilistic model, it also decomposes into two very intuitive objectives, with the first corresponding to unsupervised representation learning and the second corresponding to reinforcement learning. 

\subsection{Implementation Details}
\label{subsection:implementation}
To optimize the above objective, we extend soft actor-critic (SAC)~\cite{haarnoja2018soft}, which implements maximum entropy off-policy RL. We introduce an inference network that outputs a distribution over latent variables for the $i$-th episode, $q(\zi | \tau^i)$, conditioned on the trajectory from the $i$-th episode. The inference network, parameterized as a feedforward neural network, outputs parameters of a Gaussian distribution, and we use the reparameterization trick~\cite{kingma2013auto} to sample  $\z$. A decoder neural network reconstructs transitions and rewards given the latent embedding $\zi$, current state $\st$, and action taken $\at$, i.e. $p(\stp | \st, \at; \zi)$ and $p(r_t | \st, \at; \zi)$. Finally, $p(\zi | \zim)$ and $p(\zi | \tau^{1:i-1})$ are approximated with a shared long short-term memory (LSTM) network~\cite{hochreiter1997long}, which, at each episode $i$, receives $\zim$ from $q(\zim | \tau^{i-1})$ and hidden state $h_{i-1}$, and produces $\zi$ and the next hidden state $h_i$. 

\begin{figure}
    \centering
    \begin{minipage}{.55\textwidth}
        \begin{algorithm}[H]
            \caption{Lifelong Latent Actor-Critic (LILAC)}
            \label{alg:lilac}
            \begin{algorithmic}
            \STATE {\bfseries Input:} env, $\alpha_Q$, $\alpha_\pi$, $\alpha_\phi$, $\alpha_\text{dec}$, $\alpha_p$ \\
            \STATE Randomly initialize $\theta_Q$, $\theta_\pi$, $\phi$, $\theta_\text{dec}$, and $\theta_p$
            \STATE Initialize empty replay buffer $\mathcal{D}$
            \STATE Assign $\z^0 \gets \vec{0} $
            \FOR {i = 1, 2, \dots}
                \STATE Sample $\z^i \sim p_{\psi}(\z^i | \z^{i-1})$
                \STATE Collect trajectory $\tau^i$ with $\pi_\theta(\ac | \s, \z)$
                \STATE Update replay buffer $\mathcal{D}[i] \gets \tau^i$
                \FOR {j = 1, 2, \dots, N}
                    \STATE Sample a batch of episodes $E$ from $\mathcal{D}$
                    \STATE $\triangleright$ Update actor and critic
                    \STATE $\theta_Q \gets \theta_Q - \alpha_Q \nabla_{\theta_Q} \mathcal{J}_Q$
                    \STATE $\theta_\pi \gets \theta_\pi - \alpha_\pi \nabla_{\theta_\pi} \mathcal{J}_\pi$
                    \STATE $\triangleright$ Update inference network
                    \STATE $\phi \gets \phi - \alpha_\phi \nabla_{\phi} \left( \mathcal{J}_\text{dec} + \mathcal{J}_\text{KL} + \mathcal{J}_Q \right)$
                    \STATE $\triangleright$ Update model
                    \STATE $\theta_\text{dec} \gets \theta_\text{dec} - \alpha_\text{dec} \nabla_{\theta_\text{dec}} \mathcal{J}_\text{dec}$ 
                    \STATE $\theta_p \gets \theta_p - \alpha_p \nabla_{\theta_p} \mathcal{J}_\text{KL}$
                \ENDFOR
            \ENDFOR
            \end{algorithmic}
        \end{algorithm}
    \end{minipage}
    \hspace{0.02\textwidth}
    \begin{minipage}{.4\textwidth}
        \centering
        \includegraphics[width=\linewidth]{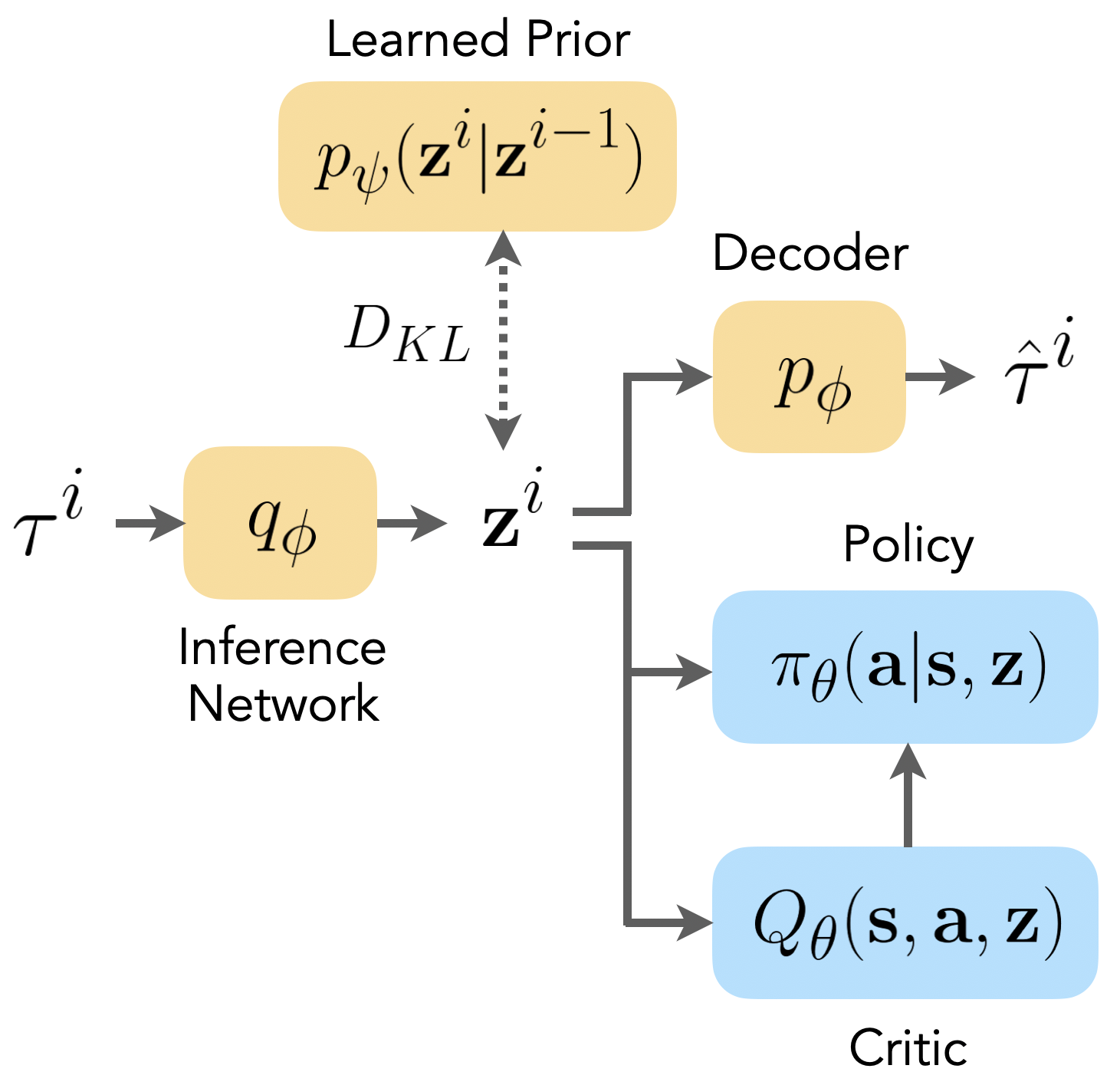}
        \caption{\footnotesize An overview of our network architecture. Our method consists of the actor $\pi$, the critic $Q$, an inference network $q$, a decoder network, and a learned prior over latent embeddings. Each component is implemented with a neural network.}
        \label{fig:network_diagram}
    \end{minipage}
\end{figure}

We visualize the entire network at a high level and how the different components interact in Figure~\ref{fig:network_diagram}. 
As depicted, the policy and critic are both conditioned on the environment state and the latent variables $\z$. During training, $\z$ is sampled from $q(\zi | \tau^i)$ outputted by the inference network. At execution time, the latent variables $\z$ the policy receives are given by the LSTM network, based on the inferred latent variables from the previous episode. Following SAC~\cite{haarnoja2018soft}, the actor loss $\mathcal{J}_\pi$ and critic loss $\mathcal{J}_Q$ are
\begin{align*}
    \mathcal{J}_\pi = \mathop{\mathbb{E}}_{\substack{\tau \sim \mathcal{D} \\ \z \sim q(\cdot | \tau) }} \left[  D_\text{KL} \left( \pi(\ac | \s) \bigg|\bigg| \frac{\exp(Q(\s, \ac, \z))}{Z(\st)} \right) \right], \hspace{0.3cm} \mathcal{J}_Q = \mathop{\mathbb{E}}_{\substack{\tau \sim \mathcal{D} \\ \z \sim q(\cdot | \tau)}} [ (Q(\s, \ac, \z) - (r + V(\s', \z)))^2 ],
\end{align*}
where $V$ denotes the target network.
Our complete algorithm, Lifelong Latent Actor-Critic (LILAC), is summarized in Algorithm~\ref{alg:lilac}.

\section{Related Work}
\label{sec:related_work}

\textbf{Partial observability in RL}. The POMDP is a general, flexible framework capturing non-stationarity and partial observability in sequential decision-making problems. While exact solution methods are tractable only for tiny state and actions spaces~\cite{kaelbling1998planning}, methods based (primarily) on approximate Bayesian inference have enabled scaling to larger problems over the course of the past two decades~\cite{kurniawati2008sarsop, roy2005finding}. In recent years, representation learning, and especially deep learning paired with amortized variational inference, has enabled scaling to a larger class of problems, including continuous state and action spaces~\cite{igl2018deep,han2019variational,lee2019stochastic, hafner2019learning} and image observations~\cite{lee2019stochastic,kapturowski2018recurrent}. However, the generality of the POMDP formulation both ignores possible performance improvements that may be realized by exploiting the structure of the DP-MDP, and does not explicitly consider between-episode non-stationarity. 

A variety of intermediate problem statements between episodic MDPs and POMDPs have been proposed. The Bayes-adaptive MDP formulation (BAMDP)~\cite{duff2002optimal, ross2008bayes}, as well as the hidden parameter MDP (HiP-MDP)~\cite{doshi2016hidden} consider an MDP with unknown parameters governing the reward and dynamics, which we aim to infer online over the course of one episode. In this formulation, the exploration-exploitation dilemma is resolved by augmenting the state space with a representation of posterior belief over the latent parameters. As noted by~\citet{duff2002optimal} in the RL literature and~\citet{feldbaum1960dual,bar1974dual} in control theory, this representation rapidly becomes intractable due to exploding state dimensionality. Recent work has developed effective methods for policy optimization in BAMDPs via, primarily, amortized inference~\cite{zintgraf2019varibad, rakelly2019efficient, lee2018bayesian}. However, the BAMDP framework does not address the dynamics of the latent parameter between episodes, assuming a temporally-fixed structure. In contrast, we are capable of modeling the evolution of the latent variable over the course of episodes, leading to better priors for online inference. 

A strongly related setting is the hidden-mode MDP~\cite{choi2000hidden}, which augments the MDP with a latent parameter that evolves via a hidden Markov model with a discrete number of states. In both the HM-MDP and the DP-MDP, the latent variable evolves infrequently, as opposed to at every time step as in the POMDP. The HM-MDP is limited to a fixed number of latent variable states due to the use of standard HMM inference algorithms. In contrast, our approach allows continuous latent variables, thus widely extending the range of applicability. 

\textbf{Non-stationarity in learning}. LILAC also shares conceptual similarities with methods from online learning and lifelong learning~\cite{ShaiBook,gama2014survey}, which aim to capture non-stationarity in supervised learning, as well as meta-learning and meta-reinforcement learning algorithms, which aim to rapidly adapt to new settings. 
Within meta-reinforcement learning, two dominant techniques exist: optimization-based~\cite{finn2017model,rothfuss2018promp,zintgraf2018fast, stadie2018some} and context-based, which includes both recurrent architectures~\cite{duan2016rl, wang2016learning, mishra2017simple} and architectures based on latent variable inference~\cite{rakelly2019efficient, lee2019stochastic, zintgraf2019varibad}. LILAC fits into this last category within this taxonomy, but extends previous methods by considering inter-episode latent variable dynamics. Previous embedding-based meta-RL algorithms---while able to perform online inference of latent variables and incorporate this posterior belief into action selection---do not consider how these latent variables evolve over the lifetime of the agent, as in the DP-MDP setting. 
The inner latent variable inference component of LILAC possesses strong similarities to the continual and lifelong learning setting~\cite{gama2014survey}. 
Many continual and lifelong learning aim to learn a variety of tasks without forgetting previous tasks~\cite{kirkpatrick2017overcoming,zenke2017continual,gradient_episodic,aljundi2019task,parisi2019continual,rusu2016progressive,shmelkov2017incremental,rebuffi2017icarl,shin2017continual}. We consider a setting where it is practical to store past experiences in a replay buffer~\cite{rolnick2019experience,finn2019online}. Unlike these prior works, LILAC aims to learn the dynamics associated with latent factors, and perform online inference. 

Concurrent work from~\citet{chandak2020optimizing} studies a setting similar to ours, where the reward and transition dynamics change smoothly across episodes, and proposes to use curve-fitting to estimate performance on future MDPs and learns a single policy that optimizes for future performance. This need for continual policy adaptation can result in performance lag in quickly changing environments; in contrast, LILAC learns a latent variable-conditioned policy, where different MDPs map to different values for these latent variables, and thus should be less sensitive to the rate of non-stationarity.

\begin{figure*}
    \centering
    \includegraphics[width=0.95\linewidth]{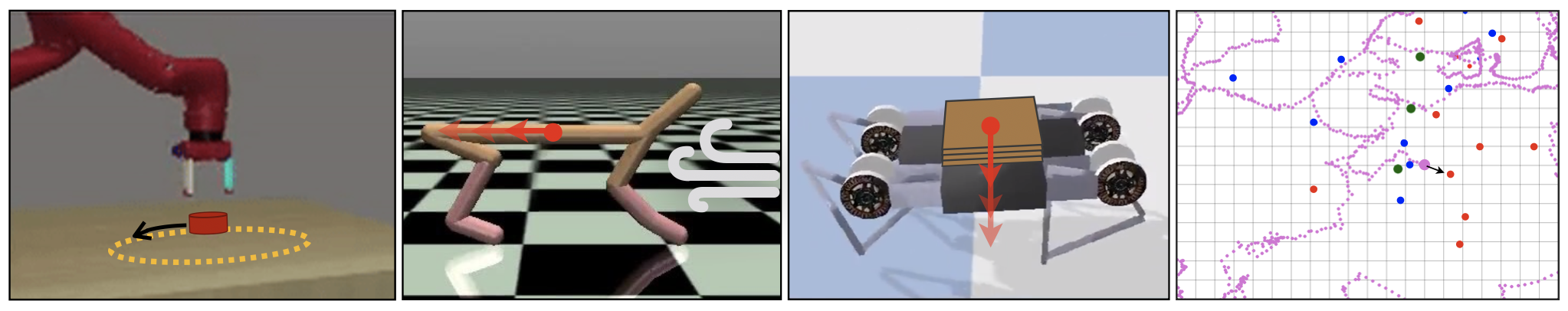}
    \caption{\footnotesize The environments in our evaluation. Each environment changes over the course of learning, including a changing target reaching position (left), variable wind and goal velocities (middle left), and variable payloads (middle right). We also introduce a 2D open world environment with non-stationary dynamics and visualize a partial snapshot of the LILAC agent's lifetime in purple (right).}
    \label{fig:environments}
    \vspace{-0.4cm}
\end{figure*}

\section{Experiments}
\label{sec:experiments}
In our experiments, we aim to address our central hypothesis: that existing off-policy RL algorithms struggle under persistent non-stationarity and that, by leveraging our latent variable model, our approach can make learning in such settings both effective and efficient.

\textbf{Environments.} We construct four continuous control environments with varying sources of change in the reward and/or dynamics. These environments are designed such that the policy needs to change in order to achieve good performance. 
The first is derived from the simulated Sawyer reaching task in the Meta-World benchmark~\cite{yu2019meta}, in which the target position is not observed and moves between episodes. In the second environment based on Half-Cheetah from OpenAI Gym~\cite{brockman2016openai}, we consider changes in the direction and magnitude of wind forces on the agent, and changes in the target velocity. We next consider the 8-DoF minitaur environment~\cite{tan2018sim} and vary the mass of the agent between episodes, representative of a varying payload. Finally, we construct a 2D navigation task in an \textit{infinite, non-episodic} environment with non-stationary dynamics which we call 2D Open World. The agent's goal is to collect food pellets and to avoid other objects and obstacles, whilst subjected to unknown perturbations that vary on an episodic schedule. These environments are illustrated in Figure~\ref{fig:environments}. For full environment details, see Appendix~\ref{app:envs}.

\textbf{Comparisons.} 
We compare our approach to standard soft-actor critic (SAC)~\cite{haarnoja2018soft}, which corresponds to our method without any latent variables, allowing us to evaluate the performance of off-policy algorithms amid non-stationarity. We also compare to stochastic latent actor-critic (SLAC)~\cite{lee2019stochastic}, which learns to model partially observed environments with a latent variable model but does not address inter-episode non-stationarity. This comparison allows us to evaluate the importance of modeling non-stationarity between episodes. Finally, we include proximal policy optimization (PPO)~\cite{schulman2017proximal} as a comparison to on-policy RL. Since the tasks in the Sawyer and Half-Cheetah domains involve goal reaching, we can obtain an oracle by training a goal-conditioned SAC policy, i.e. with the true goal concatenated to the observation. We provide this comparison to help contextualize the performance of our method against other algorithms. We tune hyperparameters for all approaches, and run each with the best hyperparameter setting with 3 random seeds. For all hyperparameter details, see Appendix~\ref{app:hyperparameters}.

\begin{figure*}
    \centering
    \includegraphics[width=0.32\linewidth]{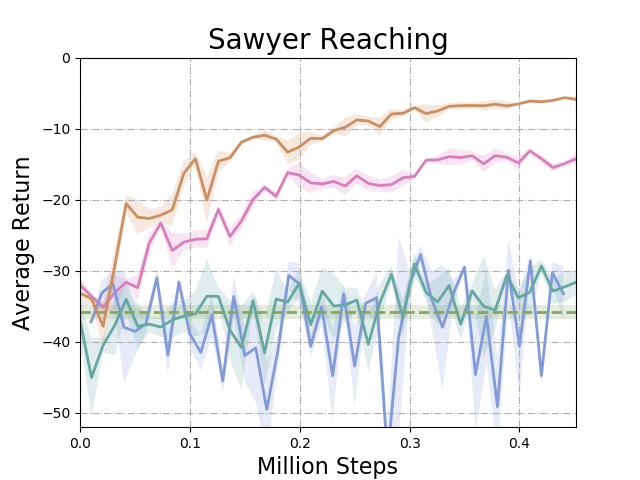}
    \includegraphics[width=0.32\linewidth]{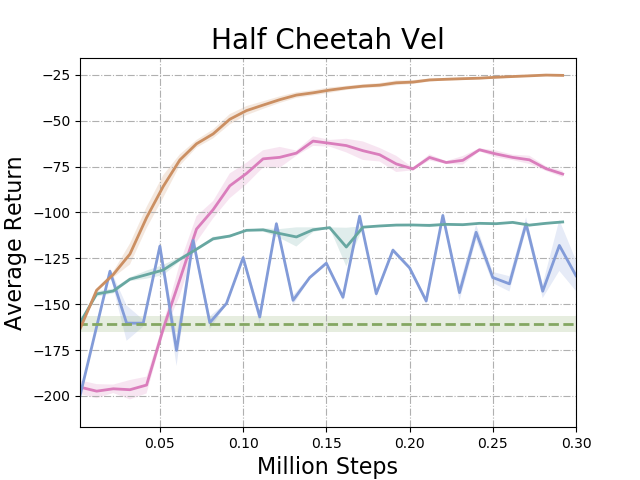}
    \includegraphics[width=0.32\linewidth]{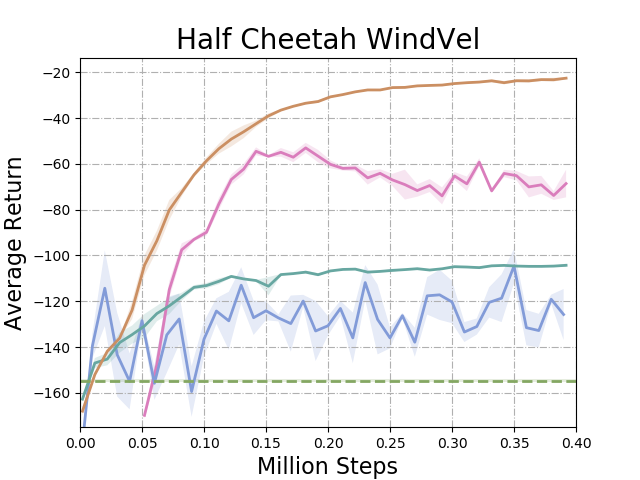} \\
    \includegraphics[width=0.32\linewidth]{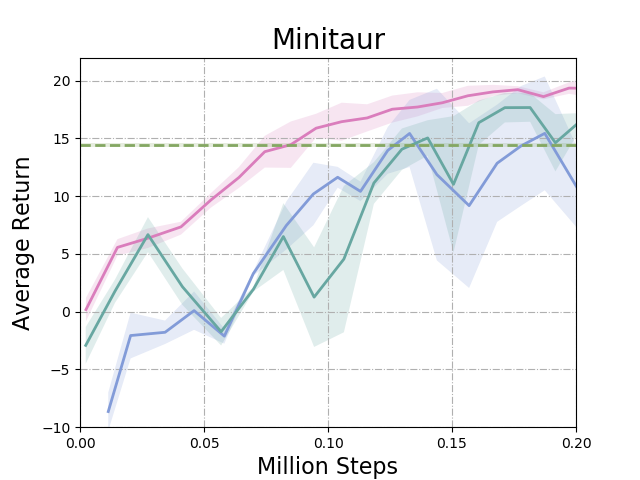}
    \includegraphics[width=0.32\linewidth]{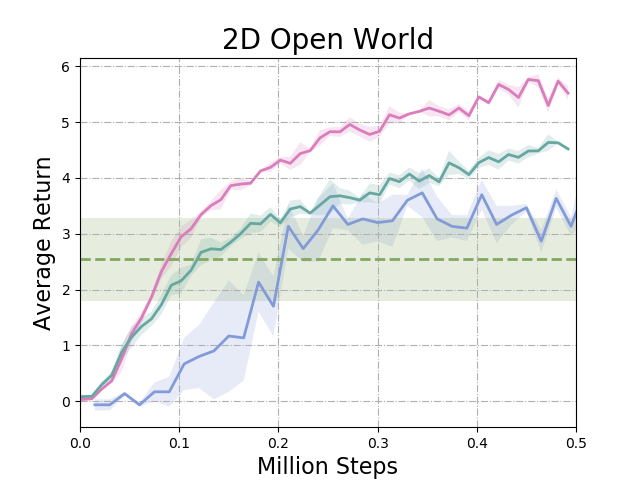} \\
    \vspace{0.1cm}
    \includegraphics[width=0.5\linewidth]{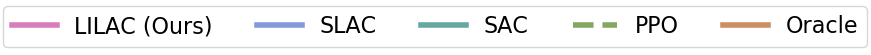}
    \caption{\footnotesize Learning curves across our experimental domains. In all settings, our approach is substantially more stable and successful than SAC, SLAC, and PPO. As demonstrated in Half-Cheetah with varying target velocities and wind forces, our method can cope with non-stationarity in \textit{both} dynamics and rewards.}
    \label{fig:experiments}
    \vspace{-0.3cm}
\end{figure*}

\textbf{Results.} Our experimental results are shown in Figure~\ref{fig:experiments}. Since on-policy algorithms tend to have worse sample complexity, we run PPO for 10 million environment steps and plot only the asymptotic returns. In all domains, LILAC attains higher and more stable returns compared to SAC, SLAC, and PPO. Since SAC amortizes experience collected across episodes into a single replay buffer, we observe that the algorithm converges to an averaged behavior. Meanwhile, SLAC does not have the mechanism to model non-stationarity across episodes, and has to infer the unknown dynamics and reward from the initial steps taken during each episode, which the algorithm is not very successful at. Due to the cyclical nature of the tasks, the learned behavior of SLAC results in oscillating returns across tasks. Similarly, PPO cannot adapt to per-episode changes in the environment and ultimately converges to learning an average policy. In contrast to these methods, LILAC infers how the environment changes in future episodes and steadily maintains high rewards over the training procedure, despite experiencing persistent shifts in the environment in each episode. Further, LILAC can learn under simultaneous shifts in \textit{both} dynamics and rewards, verified by the HC WindVel results. LILAC can also adeptly handle shifts in the 2D Open World environment without episodic resets. A partial snapshot of the agent's lifetime from this task is visualized in Figure~\ref{fig:environments}.

\begin{wrapfigure}{r}{0.33\textwidth}
    \centering
    \vspace{-0.7cm}
    \includegraphics[width=\linewidth]{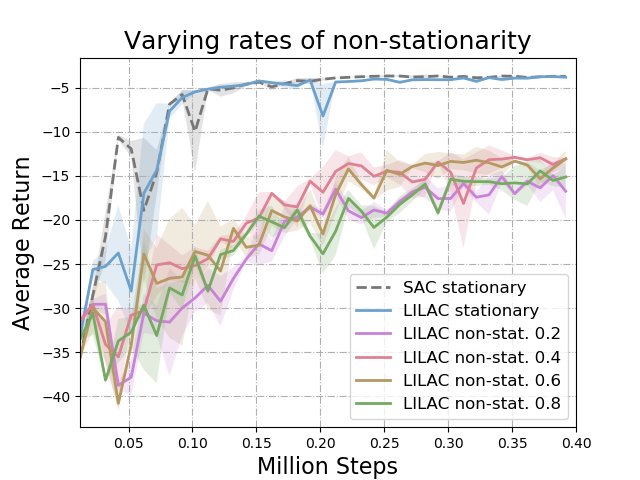}
    \vspace{-0.6cm}
    \caption{\footnotesize We evaluate LILAC in the Sawyer task with varying rates of non-stationarity by moving the goal $0.2$, $0.4$, $0.6$, and $0.8$ radians along a fixed-radius circle between episodes. We also plot the performance of LILAC under stationary conditions (with the goal fixed).}
    \label{fig:rate_results}
    \vspace{-0.7cm}
\end{wrapfigure}
\textbf{Rate of environment shift.} We next evaluate whether LILAC can handle varying rates of non-stationarity. To do so, we use the Sawyer reaching domain, where the goal moves along a fixed-radius circle, and vary the step size along the circle ($0.2, 0.4, 0.6$, and $0.8$ radians/step) to generate environments that shift at different speeds. As depicted in Figure~\ref{fig:rate_results}, LILAC's performance is largely independent of the environment's rate of change. We also evaluate LILAC under stationary conditions, i.e. with a fixed goal, and find that LILAC achieves the same performance as SAC, thus retaining the ability to learn as effectively as SAC in a fixed environment. These results demonstrate LILAC's efficacy under a range of rates of non-stationarity, including the stationary case.

The gap in LILAC's performance between the non-stationary and stationary cases is likely due to imprecise estimates of future environment conditions given by the prior $p_\phi(\z' | \z)$. Currently, the executed policy uses a fixed $\z$ given by the prior for the entire duration of the episode, but a natural extension that may improve performance is updating $\z$ \textit{during} each episode. In particular, we could encode the collected partial trajectory with the inference network and combine the inferred values with the prior to form an updated estimate, akin to Bayesian filtering.

\section{Conclusion}
\label{sec:conclusion}

We considered the problem of reinforcement learning with lifelong non-stationarity, a problem which we believe is critical to reinforcement learning systems operating in the real world. This problem is at the intersection of reinforcement learning under partial observability (i.e. POMDPs) and online learning; hence we formalized the problem as a special case of a POMDP that is also significantly more tractable. We derive a graphical model underlying this problem setting, and utilize it to derive our approach under the formalism of reinforcement learning as probabilistic inference~\cite{levine2018reinforcement}. Our method leverages this latent variable model to model the change in the environment, and conditions the policy and critic on the inferred values of these latent variables. On a variety of challenging continuous control tasks with significant non-stationarity, we observe that our approach leads to substantial improvement compared to state-of-the-art reinforcement learning methods.

While the DP-MDP formulation represents a strict generalization of the commonly-considered meta-reinforcement learning settings (typically, a BAMDP~\cite{zintgraf2019varibad}), it is still somewhat limited in its generality. In particular, the assumption of task parameters shifting \textit{between} episodes, but never during them, presents a possibly unrealistic limitation. While relaxing this assumption leads, in the worst case, to a POMDP, there is potentially additional structure that may be exploited under the HM-MDP~\cite{choi2000hidden} assumption of infrequent, discrete, unobserved shifts in the task parameters. In particular, this notion of infrequent, discrete shifts underlies the changepoint detection literature~\cite{adams2007bayesian, fearnhead2007line}. Previous work both within sequential decision making in changing environments~\cite{da2006dealing,hadoux2014sequential,banerjee2017quickest} and meta-learning within changing data streams~\cite{harrison2019continuous} may enable a version of LILAC capable of handling unobserved changepoints, and this setting is likely a fruitful direction for future research.

\begin{ack}
The authors would like to thank Allan Zhou, Evan Liu, and Laura Smith for helpful feedback on an early version of this paper. Annie Xie was supported by an NSF fellowship. Chelsea Finn is a CIFAR fellow.
\end{ack}

\bibliography{bibtex}
\bibliographystyle{plainnat}

\clearpage
\section*{Appendix}
\renewcommand{\thesubsection}{\Alph{subsection}}

\subsection{Environment Details}
\label{app:envs}
Below, we provide environment details for each of the four experimental domains.

\subsubsection{Sawyer Reaching}
In this environment, which is based on the simulated Sawyer reaching task in the Meta-World suite~\cite{yu2019meta}, the goal is to reach a particular position. The target position, which is unobserved throughout, moves after each episode.

The episodes are $150$ timesteps long, and the state is the position of the end-effector in $( x, y, z )$ coordinate space. The actions correspond to changes in end-effector positions. The reward is defined as
$$
r(\s, \ac) = - \| \s - \s_g \|_2,
$$
where $\s_g$ at episode $i$ is defined as
$$
\s_g = \begin{bmatrix}
    0.1 \cdot \cos(0.5 \cdot i) \\
    0.1 \cdot \sin(0.5 \cdot i) \\
    0.2
\end{bmatrix}.
$$
In other words, the sequence of goals is defined by a circle in the $xy$-plane. For the oracle comparison, the sequence of goals and the reward function are the same, except the state observation here is the concatenation of the end-effector position and the goal position $\s_g$.

\subsubsection{Half-Cheetah Vel}
This environment builds off of the Half-Cheetah environment from OpenAI Gym~\cite{brockman2016openai}. In this domain, the agent must reach a target velocity in the $x$-direction, which varies across episodes, i.e., the reward is
$$
r(\s, \ac) = -\| v_s - v_g \|_2 - 0.05 \cdot \| \ac \|_2,
$$
where $v_s$ is the observed velocity of the agent. The state consists of the position and velocity of the agent's center of mass and the angular position and angular velocity of each of its six joints, and actions correspond to torques applied to each of the six joints. 

The target velocity $v_g$ varies according to a sine function, i.e., the target velocity for episode $i$ is
$$
v_g = 1.5 + 1.5 \sin (0.5 \cdot i).
$$
For the oracle comparison, the target velocity $v_g$ is appended to the state observation. Each episode, across all comparisons, is $50$ timesteps long.

\subsubsection{Half-Cheetah Wind+Vel}
In this variant of Half-Cheetah Vel, the agent is additionally subjected to varying wind forces. The force for each episode is defined by 
$$
f_w = 10 + 10 \sin(0.2 \cdot i)
$$
and is applied constantly along the $x$-direction throughout the episode.

\subsubsection{Minitaur Mass}
We use the simulated Minitaur environment developed by~\citet{tan2018sim}. We induce non-stationarity by varying the mass of the agent between episodes akin to increasing and decreasing payloads. Specifically, the mass at each episode is
$$
m = 1.0 + 0.75 \sin(0.3 \cdot i).
$$
The reward is defined by
$$
r(\s_t, \ac_t) = 0.3 - |0.3 - \s_{t,v}| - 0.01 \cdot \|\at - 2 \ac_{t-1} + \ac_{t-2}\|_1,
$$
where the first two terms correspond to the velocity reward, which encourages the agent to run close to a target velocity of $0.3$ m/s, and the last term corresponds to an acceleration penalty defined by the last three actions taken by the agent. The state includes the angles, velocities, and torques of all eight motors, and the action is the target motor angle for each motor. Each episode is $100$ timesteps long.

\subsubsection{2D Open World}
Finally, we design an infinite, non-episodic environment in which the agent's goal is to collect red pellets and avoid blue pellets and other obstacles. Simultaneously, the agent is subjected non-stationary dynamics, in particular directed wind forces $f_w$ and action re-scaling $c$. These dynamics change after every $100$ timesteps, which is known to the agent, according to the following equations:
\begin{align*}
    f_w &= \begin{bmatrix}
        0.015 \cos(0.2 \cdot \left\lfloor t/100 \right\rfloor) \\
        0.015 \sin(0.2 \cdot \left\lfloor t/100 \right\rfloor)
    \end{bmatrix} \\
    c &= 0.03 + 0.015 \sin (0.125 \cdot \left\lfloor t/100 \right\rfloor)
\end{align*}

\subsection{Hyperparameter Details}
\label{app:hyperparameters}
In this section, we provide the hyperparameter values used for each method. 

\subsubsection{LILAC (Ours)}
\textit{Latent space.} For our method, we use a latent space size of $8$ in Sawyer Reaching and 2D Open World, and size of $40$ in the other experiments: Half-Cheetah Vel, Half-Cheetah Wind+Vel, and Minitaur Mass. 

\textit{Inference and decoder networks.} The inference and decoder networks are MLPs with 2 fully-connected layers of size $64$ in Sawyer Reaching and 2D Open World; 1 fully-connected layer of size $512$ in Half-Cheetah Vel and Half-Cheetah Wind+Vel; and 2 fully-connected layers of size $512$ in Minitaur Mass. 

\textit{Policy and critic networks.} The policy and critic networks are MLPs with 3 fully-connected layers of size $256$ in the Sawyer Reaching experiment; and 2 fully-connected layers of size $256$ in the other experiments.


For the Sawyer Reaching experiment, $\beta_1$ and $\beta_2$ are
\begin{align*}
    \beta_1 &= \begin{cases}
        0, & \text{iter} < 10000 \\
        1, & \text{iter} \ge 10000
    \end{cases} \\
    \beta_2 &= \begin{cases}
        0, & \text{iter} < 10000 \\
        \min(1\text{e-}6 \cdot (\text{iter} - 10000), 1 ), & \text{iter} \ge 10000
    \end{cases}
\end{align*}

For Half-Cheetah Vel and Half-Cheetah Wind+Vel, $\beta_1$ and $\beta_2$ are
\begin{align*}
    \beta_1 &= \begin{cases}
        0, & \text{iter} < 50000 \\
        1, & \text{iter} \ge 50000
    \end{cases} \\
    \beta_2 &= \begin{cases}
        0, & \text{iter} < 10000 \\
        \min(1\text{e-}6 \cdot (\text{iter} - 10000), 1 ), & \text{iter} \ge 10000
    \end{cases}
\end{align*}

For the Minitaur Mass experiment, $\beta_1$ and $\beta_2$ are
\begin{align*}
    \beta_1 &= \begin{cases}
        0, & \text{iter} < 10000 \\
        1, & \text{iter} \ge 10000
    \end{cases} \\
    \beta_2 &= \begin{cases}
        0, & \text{iter} < 10000 \\
        1\text{e-}6, & \text{iter} \ge 10000
    \end{cases}
\end{align*}
    
Finally, for 2D Open World, $\beta_1$ and $\beta_2$ are
\begin{align*}
    \beta_1 &= 0 \\
    \beta_2 &= \begin{cases}
        1\text{e-}5, & \text{iter} < 10000 \\
        \min(1\text{e-}5 \cdot (\text{iter} - 10000), 1 ), & \text{iter} \ge 10000
    \end{cases}
\end{align*}

\subsubsection{Stochastic Latent Actor-Critic}
\textit{Latent space.} SLAC factorizes its per-timestep latent variable $\z_t$ into two stochastic layers $\z_t^1$ and $\z_t^2$, i.e. $p(\z_t) = p(\z_t^2 | \z_t^1) p(\z_t^1)$. In the Sawyer Reaching and 2D Open World experiments, the size of $\z_t^1$ is $16$ and the size of $\z_t^2$ is $8$. In all other experiments, the size of $\z_t^1$ is $64$ and the size of $\z_t^2$ is $32$. 

\textit{Inference and decoder networks.} The inference and decoder networks are MLPs with 2 fully-connected layers of size $64$ in Sawyer Reaching and 2D Open World; 1 fully-connected layer of size $512$ in Half-Cheetah Vel and Half-Cheetah Wind+Vel; and 2 fully-connected layers of size $512$ in Minitaur Mass. 

\textit{Policy and critic networks.} The policy and critic networks are MLPs with 3 fully-connected layers of size $256$ in the Sawyer Reaching experiment; and 2 fully-connected layers of size $256$ in the other experiments.

\subsubsection{Soft Actor-Critic}
\textit{Policy and critic networks.} The policy and critic networks are MLPs with 3 fully-connected layers of size $256$ in the Sawyer Reaching experiment; and 2 fully-connected layers of size $256$ in the other experiments.

\end{document}